\documentclass[letterpaper, 10 pt, conference]{ieeeconf}  

\IEEEoverridecommandlockouts                              

\usepackage{graphicx}
\usepackage{amsfonts}

\title{\LARGE \bf
Occupancy Map Prediction Using Generative and Fully Convolutional Networks for Vehicle Navigation}

\author{Kapil Katyal$^{1,2}$, Katie Popek$^{1}$, Chris Paxton$^{2}$, Joseph Moore$^{1}$,\\ Kevin Wolfe$^{1}$, Philippe Burlina$^{1}$, and Gregory D.  Hager$^{2}$
\thanks{$^{1}$Research and Exploratory Development Dept., Johns Hopkins University Applied Physics Lab, Laurel, MD, USA.
        {\tt\small \{Kapil.Katyal, Katie.Popek, Joseph.Moore, Kevin.Wolfe, Phillippe.Burlina\}@jhuapl.edu}}%
\thanks{$^{2}$Dept. of Computer Science, Johns Hopkins University, Baltimore, MD, USA.
        {\tt\small \{cpaxton, ghager1\}@jhu.edu}}%
}

\def\BibTeX{{\rm B\kern-.05em{\sc i\kern-.025em b}\kern-.08em
    T\kern-.1667em\lower.7ex\hbox{E}\kern-.125emX}}
\begin{document}

\maketitle
\thispagestyle{empty}
\pagestyle{empty}

\begin{abstract}
Fast, collision-free motion through unknown environments remains a challenging problem for robotic systems. In these situations, the robot's ability to reason about its future motion is often severely limited by sensor field of view (FOV). By contrast, biological systems routinely make decisions by taking into consideration what might exist beyond their FOV based on prior experience. In this paper, we present an approach for predicting occupancy map representations of sensor data for future robot motions using deep neural networks. We evaluate several deep  network architectures, including purely generative and adversarial models. Testing on both simulated and real environments we demonstrated performance both qualitatively and quantitatively, with SSIM similarity measure up to 0.899. We showed that it is possible to make predictions about occupied space beyond the physical robot's FOV from simulated training data. In the future, this method will allow robots to navigate through unknown environments in a faster, safer manner.
\end{abstract}

\section{INTRODUCTION}
High-speed, autonomous navigation is predicated on the ability to reason about the environment for effective, collision-free path planning.  Existing approaches operate on current sensor readings to update an occupancy map corresponding to an internal representation of where obstacles exist in the environment.  These occupancy maps are then used by planning algorithms to generate a collision-free path to a target goal.  One of the limitations of this approach is that the planning horizon is limited to the  field of view (FOV) of the sensor.

On the other hand, behavioral neuroscience and biological psychology point to the potential role of prediction for navigation in animals and humans. In particular, the hypocampus appears to exhibit some neuronal structures as well as firing sequences that could support not only mapping but also predictive mapping capabilities~\cite{buckner2010role}.
Indeed, humans continuously make predictions of what to expect based on past experiences. This allows us to adjust our control policy in real time depending on how close our observations match our predictions~\cite{doi:10.1152/jn.00036.2017}.  The advantage is most evident while running along a hallway approaching a T-intersection.  Even though we cannot see the left or right paths, we generally assume the straight lines will continue and we can predict how the hallway will appear as we turn the corner.  Because of this prediction, we do not adjust our running speed unless our prediction is inaccurate.  Following this intuition, we believe future predictions of occupancy maps can enable risk sensitive control policies for mobile and aerial vehicles. By being able to predict occupancy maps, we can enable faster navigation as the planning horizon can extend beyond the sensor's limited FOV. 

This concept is similar to image completion, a problem for which multiple solutions have been suggested in the past~\cite{Bertalmio:2000:II:344779.344972,Chan02mathematicalmodels}. We take an alternate approach, leveraging the fact that structural information from the observed geometry of the world that can help us make useful predictions about the environment.  Deep neural networks have significant advantages over other approaches when used for image completion or image generation~\cite{DBLP:journals/corr/YehCLHD16}.  One of the most promising innovations in deep learning has been the development of autoencoder networks and generative adversarial networks (GANs)~\cite{NIPS2014_5423}.  These networks use a minimax game adversarial training with opponent generative and  discriminative networks, and are capable of encoding a latent representation of  images used to generate new examples from  latent space. 

\begin{figure}[t]
	\centering
    \includegraphics[width=.95\columnwidth]{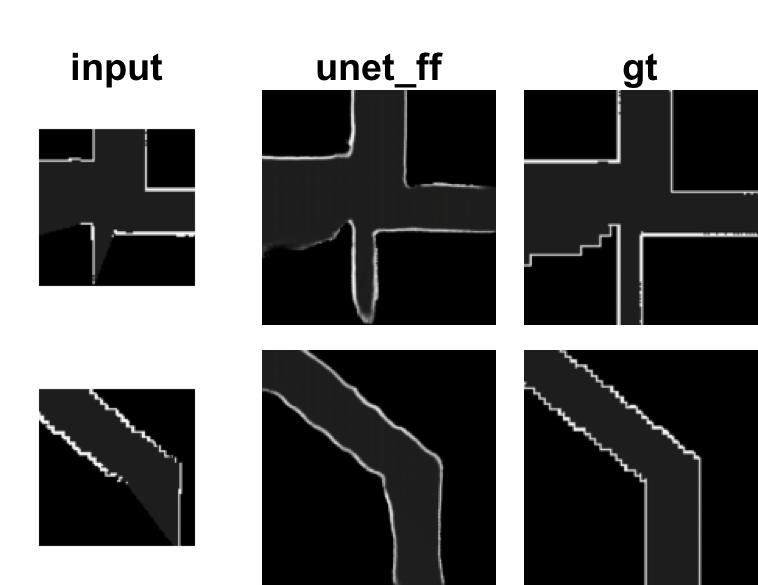}
    \caption{Two samples of predicted images.  The left image is the input based on sensor readings, middle image is the predicted expanded occupancy map (1.50x) and the right image is the ground truth.}
	\label{fig:sample_pred}
\end{figure}

In this paper, we demonstrate the ability to generate future predictions of occupancy maps without an explicit model using a variety of different neural network architectures with two examples shown in Fig.~\ref{fig:sample_pred}. 

The main contributions of this work include:
\begin{itemize}
\item A dataset consisting of simulated and physical occupancy maps that can be used to train and validate neural networks for predicting occupancy maps, 
\item A framework to evaluate the performance and accuracy of different neural network architectures,
\item Qualitative and quantitative analysis of the prediction capabilities, performance and accuracy of various network architectures, 
\item Validation of our approach using occupancy maps generated by a physical LIDAR sensor.
\end{itemize}
\section{RELATED WORK}
\subsection*{Model Predictive Control} 
High-speed navigation has been an active area of research primarily focusing on trajectory optimization, path planning and state estimation.  Several papers have investigated model predictive control (MPC) techniques for navigation including~\cite{5152468,6580446} however these approaches typically model the vehicle dynamics to predict vehicle motion and not necessarily the environment.

\subsection*{Deep Learning for Generative Models} 
Deep neural networks have been used in a number of promising ways to achieve high performance in domains such as vision, speech and more recently in robotics manipulation~\cite{finn2017deep,levine2016learning}.

Oh et. al. used feedforward and recurrent neural networks to perform action-conditional video prediction using Atari games with promising results~\cite{NIPS2015_5859}. These have also been used in image completion, e.g., by Ulyanov et al.~\cite{DBLP:journals/corr/abs-1711-10925}.  In addition,  GANs have demonstrated a promising method for image generation~\cite{NIPS2014_5423}. Isola et al. proposed an approach for training conditional GANs which create one image from another image~\cite{pix2pix2017}.

\subsection*{Deep Learning for Navigation} 

More recently, several papers have described approaches to combine elements of deep neural networks with autonomous navigation.  These include using deep neural networks for model predictive control~\cite{finn2017deep}. Tamar et al. proposed Value Iteration Networks, which embed a planner inside a deep neural net architecture~\cite{tamar2016value}. Several papers investigate the use of deep reinforcement learning to develop collision-free planning without the need of an internal map, however, these are still restricted by the sensor's FOV~\cite{DBLP:journals/corr/TaiPL17, DBLP:journals/corr/abs-1709-10489}.

While each of these papers makes promising contributions to their respective fields, none of the prior works use neural networks and in particular, GANs to generate predictions of future occupancy maps, nor do they focus on extending the planning horizon beyond the sensor's FOV.

\section{PROPOSED ARCHITECTURES}
The goal of our network architecture is to learn a function that maps an input occupancy map to an expanded occupancy map that extends beyond the FOV of the sensor.  More formally, we are learning the function
\[
	f : x \rightarrow y_i
\]
\noindent where \begin{math}x\end{math} represents the state, in this case, the input occupancy map as an image, \begin{math}y_i\end{math} represents the output occupancy map and \begin{math}i\in\mathbb{R}\end{math} represents percent increase of the expanded occupancy map.  Components of the function $f$ include an encoding function $f_{enc}(x)\rightarrow h\in \mathcal{H}$ which maps the state space, input occupancy maps to a hidden state and $f_{dec}(h) \rightarrow (y_i)$, which is a decoding function mapping the hidden state to an expanded, predicted occupancy map.

In our experiments, we compare several different neural network architectures including:

\noindent
(A) a feedforward network based on a U-Net architecture (\textbf{unet\_ff}) \\
%
(B) a feedforward network based on the ResNet architecture (\textbf{resnet\_ff}) \\
%
(C) a GAN using the feedforward network from (a) as the generative network (\textbf{gan})\\

A 5-layer multilayer perceptron was also evaluated as a baseline, however in our testing, it did not produce reliable predictions. 

\begin{figure}[t]
	\centering
    \includegraphics[width=1\columnwidth]{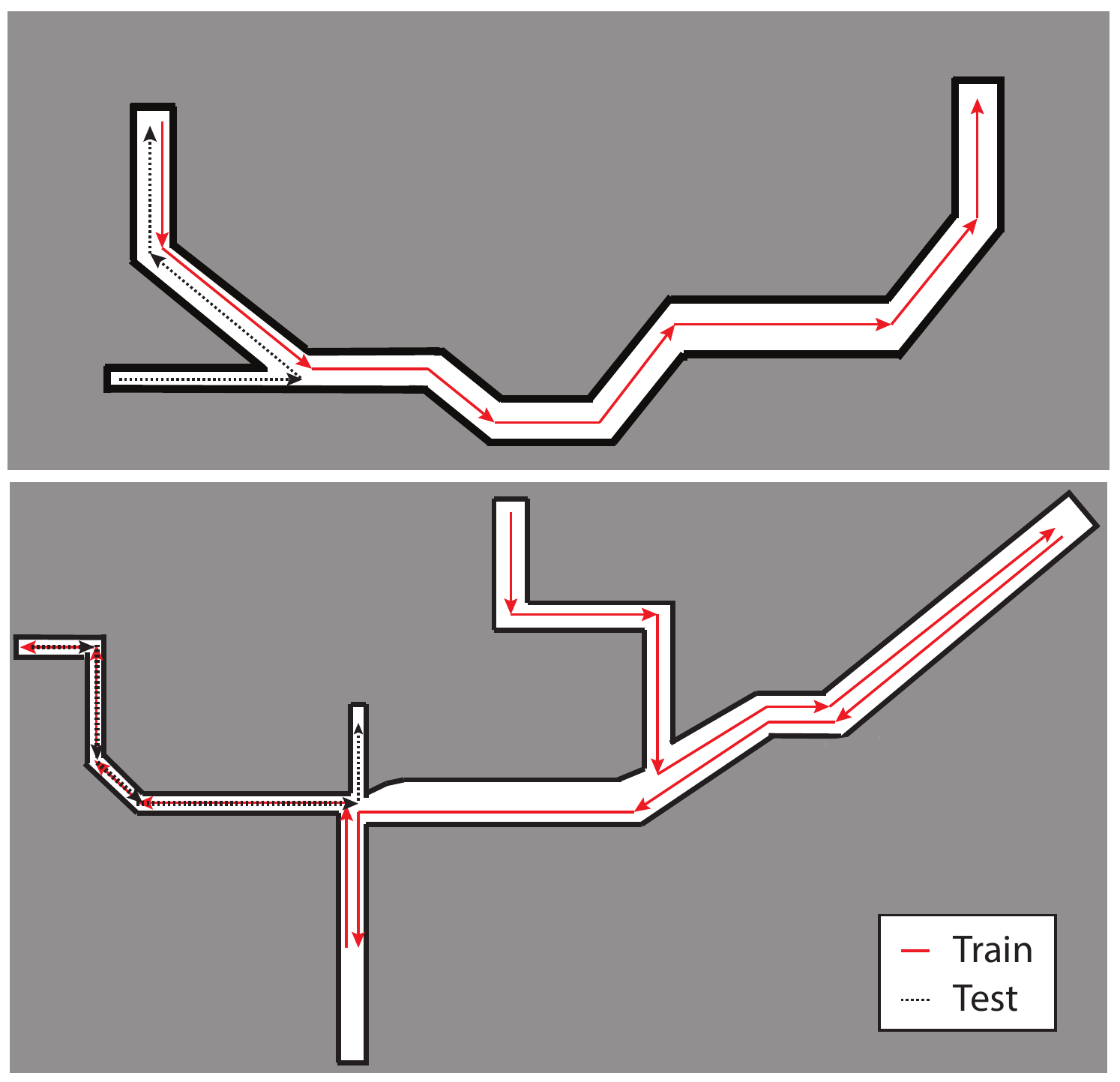}
    \caption{Simulated ground truth maps with white representing free space, black is occupied, and gray is unknown. Four trajectories were used for training (depicted in solid red), and two paths were used as the test set (dotted black).}
	\label{fig:simulated_paths}
\end{figure}

\begin{figure*}[t]
	\centering
    \includegraphics[width=.95\textwidth]{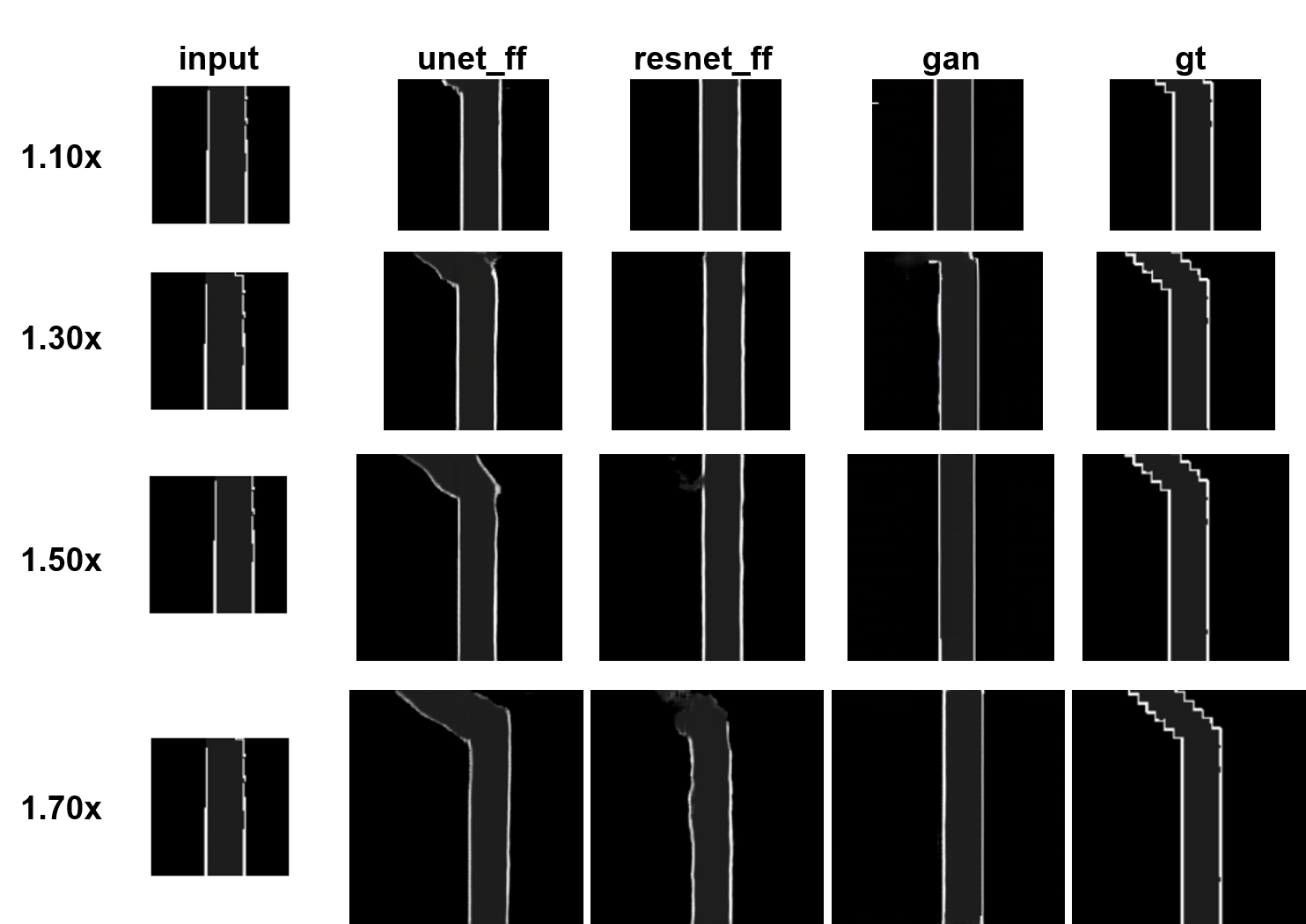}
    \caption{This figure describes the input data, the predicted images and the ground truth for each of the neural networks evaluated on the simulated dataset across an expanding prediction window from 1.10x increase to 1.70x increase.}
	\label{fig:good_sim_images}
\end{figure*}

\subsection{U-Net Feedforward Model}\label{unet}
The U-Net feedforward model is based on the network architecture defined by Ronneberger et. al~\cite{DBLP:journals/corr/RonnebergerFB15} and consists of skip connections which allows a direct connection between layers \begin{math}i\end{math} and \begin{math}n-i\end{math} enabling the option to bypass the bottleneck associated with the downsampling layers in order to perform an identity operation.  Similar to~\cite{pix2pix2017}, the encoder network consists of 8 convolution, batch normalization and ReLU layers where each convolution consists of a $4 \times 4$ filter and stride length of 2.  The number of filters for the 8 layers in the encoder network are: (64, 128, 256, 512, 512, 512, 512, 512).  The decoder network consists of 8 upsampling layers with the following number of filters: (512, 1024, 1024, 1024, 1024, 512, 256, 128).

\subsection{ResNet Feedforward Model}
The ResNet feedforward model is based on the work by Johnson et. al~\cite{DBLP:journals/corr/JohnsonAL16} which consists of 2 convolution layers with stride 2, 9 residual blocks as defined by~\cite{DBLP:journals/corr/HeZRS15} and two deconvolution layers with with a stride of \( \frac{1}{2} \).  A key reason this network was selected was based on the ability to learn identify functions, which is key to image translation as well as the success in image-to-image translation demonstrated by the CycleGAN network~\cite{DBLP:journals/corr/ZhuPIE17}. 

\subsection{GAN Model}
The GAN networks is based on the pix2pix architecture~\cite{pix2pix2017} which has demonstrated impressive results in general purpose image translation including generating street scenes, building facades and aerial images to maps. This network uses the U-Net Feedforward model defined in section ~\ref{unet} and consists of a 6 layer discriminator network with filter sizes: (64, 128, 256, 512, 512, 512).


\section{SIMULATED DATA EXPERIMENTS}

Our approach to testing occupancy map prediction using the networks defined above 
first involved generating a dataset and then performing qualitative and quantitative analysis of the predicted images compared to the ground truth. 
\subsection{Data Collection}

A dataset of approximately 6000 images of occupancy map subsets was created by simulating a non-holonomic robot moving through a two-dimensional map with a planar LIDAR sensor in C++ with ROS and the OctoMap library \cite{hornung13auro}. Two maps, shown in Fig.~\ref{fig:simulated_paths}, were created in Solidworks with the path width varying between 3.5\,m to 10\,m. These were converted into OctoMap's binary tree format using binvox \cite{binvox, nooruddin03} followed by OctoMap's binvox2bt tool. The result is an occupancy map with all unoccupied space set as free. We require space outside of the walls, shown as grey in Fig.~\ref{fig:simulated_paths}, to be marked as unknown to provide a ground truth for our estimated maps. These ground truth maps were created by fully exploring the original occupancy maps.

The robot is modeled as a Dubin’s car, with a state vector $\mathbf{x} = [x, y, \theta]$ and inputs $\mathbf{u} = [v, \dot{\theta}]$ where ($x,y$) is the robot's position, $v$ is the velocity, and  $\theta$ and $\dot{\theta}$ are the heading angle and angular velocity, respectively. For simplicity, the robot is constrained to move at fixed forward velocity of 0.5\,m/s.  A planar LIDAR sensor with a scanning area of 270$^\circ$ and range of 20\,m is used to simulate returns given the robot's current pose against the ground truth map. These simulated returns are used to create the “estimated” occupancy map. Path planning is done with nonlinear model-predictive control and direct transcription at 10\,Hz.  At each time step, a subset of the maps (both the estimated and ground truth) are saved. A 5\,m by 5\,m square centered around the robot's pose was chosen with a resolution of 0.05\,m. At each time step, the robot's current state and action space are also logged. Occupancy maps are expanded over time, so our simulation performs a continuous trajectory and the data set is built consecutively instead of randomly sampling throughout a map. A total of six trajectories were simulated. Four paths were used for training data (5221 images) and two were used as a test set (1090 images).  Ground truth datasets of the expanded occupancy maps were also generated.  These expanded occupancy maps range from 1.10x to 2.00x expansion in increments of 0.10x, e.g., a 2.00x expansion results in a 10\,m by 10\,m square subset centered around the robot.  
\subsection{Training Details}
We trained each variant of the neural network using the expanded ground truth occupancy maps from scratch for 200 epochs with a batch size of 1. A total of 15 training sessions were performed to evaluate each of the three neural network architectures across five expansion increases (1.10x, 1.30x, 1.50x, 1.70x, and 2.00x).
We use the Adam optimizer with an initial learning rate of 0.0002 and momentum parameters \begin{math}\beta_1 = 0.5, \beta_2 = 0.999\end{math}. In the feedforward models, L1 loss was used as proposed in PatchGan~\cite{pix2pix2017} and in the GAN model L1+discriminator loss was used.  The decoder layers of the network used a dropout rate of 0.50 and weights were initialized from a Normal distribution ($\mu=0, \sigma=0.2$).  All models were implemented using PyTorch~\cite{paszke2017automatic}.

 \begin{figure}[t]
 	\centering
     \includegraphics[width=.75\columnwidth]{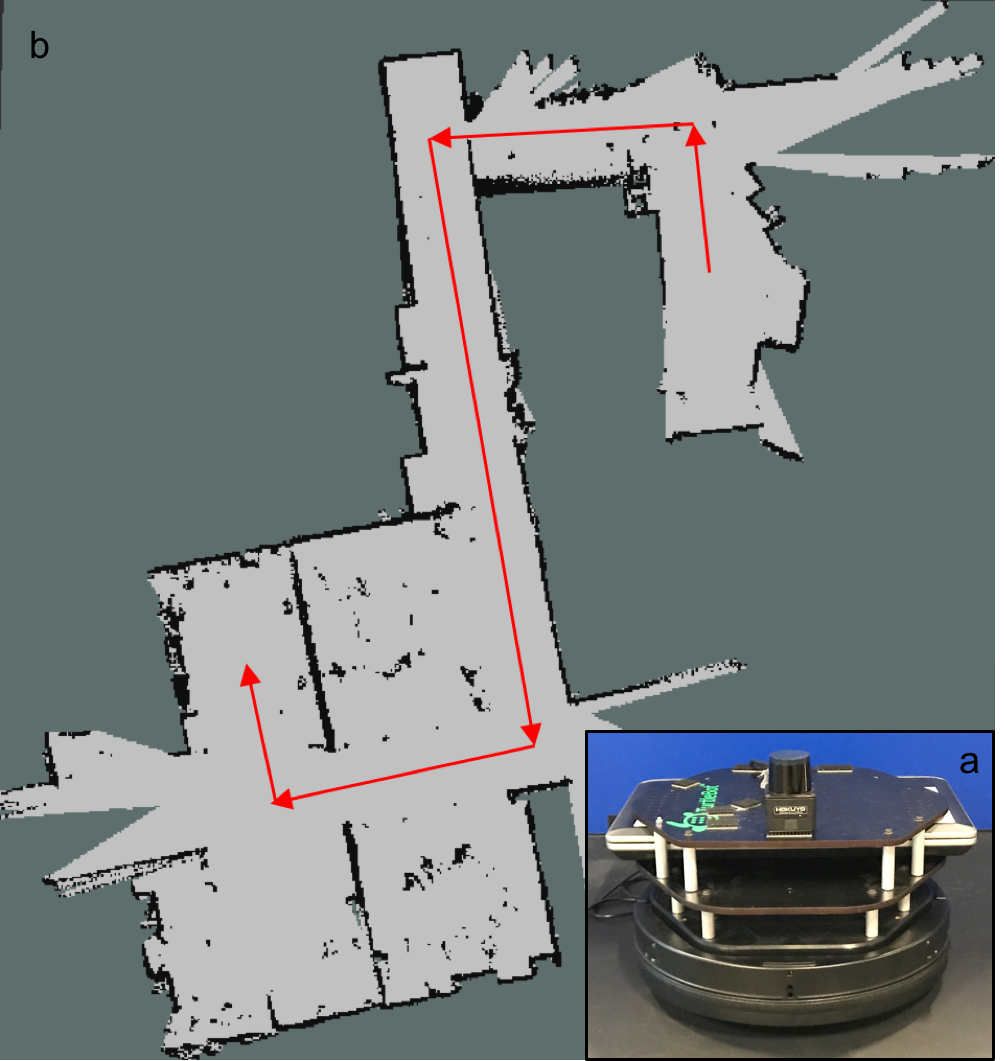}
     \caption{a). TurtleBot2 robot used to validate physical experiments. The robot has an top-mounted LIDAR sensor used to compute occupancy maps for navigation in unknown environments. b) Occupancy map created with hardware from (a). Red arrows show the robot's path.}
 	\label{fig:turtlebot_map}
 \end{figure}

\subsection{Simulation Results}
We evaluated the performance of each neural network architecture across a span of five increasing occupancy map predictions. Fig.~\ref{fig:good_sim_images} provides a snapshot of the qualitative assessment of the predicted images for each of the neural networks.  This example was selected because it demonstrates that even with  very little information, the U-Net feedforward model was able to accurately predict the presence of the surrounding obstacles while the other networks were unable to detect it.  Table~\ref{table:sa} provides the structural similarity index metric (SSIM) for each of the networks.  Based on the SSIM metric, it can be seen that the U-Net feedforward model outperforms the other networks at 1.10x and 1.30x expansion confirming the qualitative assessment.  The quality of the prediction generally decreases as the expansion percentage increases and with expansions 1.50x and above the three networks achieve similar performance.



\begin{table}[h]
\centering
\caption{SSIM Analysis for Simulation Data}
\label{table:sa}
\begin{tabular}{|c|c|c|c|} 
\hline
\textbf{Expansion} & \textbf{unet\_ff} & \textbf{resnet\_ff} & \textbf{gan} \\ [0.05ex] 
\hline\hline
1.10x & 0.899 & 0.861 & 0.818\\
\hline
1.30x & 0.818 & 0.780 & 0.790 \\
\hline
1.50x & 0.770 & 0.773 & 0.759 \\
\hline
1.70x & 0.760 & 0.752 & 0.736 \\
\hline
2.00x & 0.767 & 0.770 & 0.574 \\
\hline
\end{tabular}
\end{table}

\section{PHYSICAL EXPERIMENTS}

\begin{figure*}[t]
	\centering
    \includegraphics[width=.95\textwidth]{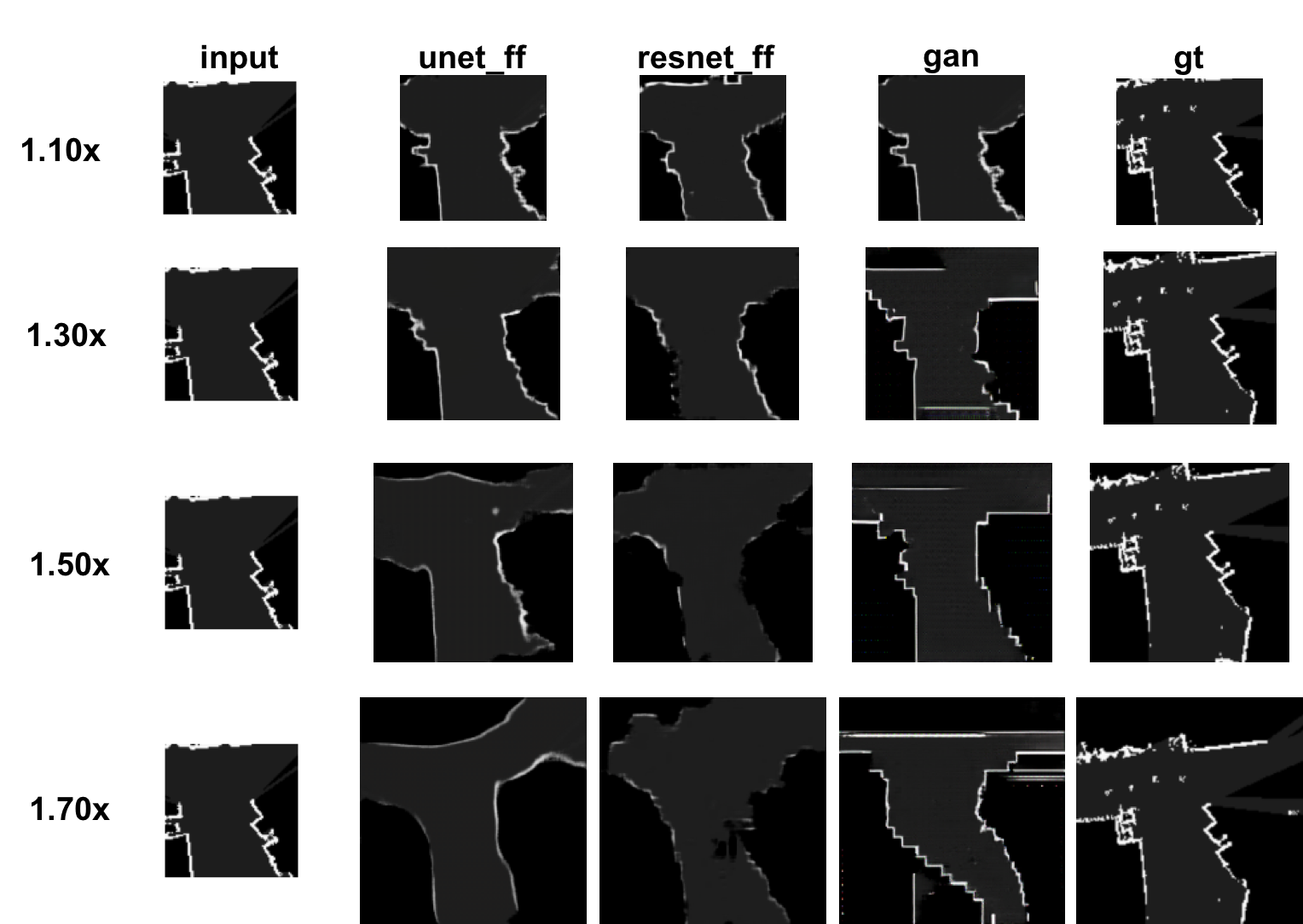}
    \caption{This figure describes the input data, the predicted images and the ground truth for each of the neural networks evaluated on the physical dataset across an expanding prediction window from 1.10x increase to 1.70x increase.}
	\label{fig:good_phys_images}
\end{figure*}
Our next experiment focused on validating our approach with occupancy maps generated by a physical LIDAR sensor. 

\subsection{Data Collection and Training Details}

In this experiment, we teleoperated a TurtleBot2 robot with a mounted Hokuyo UST-20LX LIDAR sensor (shown in Fig.~\ref{fig:turtlebot_map}(a)) around a building. The OctoMap library \cite{hornung13auro} along with a custom C++ implementation of a particle filter running at 20\,Hz was used for simultaneous localization and mapping. The final map was used as ground truth (shown in Fig.~\ref{fig:turtlebot_map}(b)). At each time step a 5\,m by 5\,m square subset centered around the robot's current pose of both the ground truth and estimated maps was saved (100 images). Expanded ground truth occupancy maps were generated ranging from 1.10x to 2.00x in 0.10x increments.

Our objective was to evaluate whether training performed on a simulated dataset could be directly transferred to occupancy maps generated by a physical LIDAR sensor.  For this reason, we opted to not fine tune the networks using the physical dataset.

\subsection{Physical Results}


Fig.~\ref{fig:good_phys_images} represents sample predictions obtained by running the networks trained using simulation data on the occupancy maps generated by the physical sensor. Table~\ref{table:sa_phys} displays the SSIM metric across each of the networks. In the physical experiments, the data is more inconclusive.  Similar to the simulation experiments, the quality of the predictions generally decrease as the predicted distance increases, however there was no noticeable difference across the three networks.

\section{DISCUSSION}

The ability to perform predictions is key to navigation. This capability is also motivated from the perspective of behavioral neuroscience and psychology.  In particular it has been found~\cite{buckner2010role} that certain neuronal structures point to mapping capabilities and may be involved in encoding predictive mapping events based on past experience. The net product is that neurons do not activate solely based on current visual input, but also based on a sequence of locations, so as to enable prediction  (see ~\cite{buckner2010role}). In this paper, our goal is to develop techniques that enable future predictions of occupied space for robotic navigation. 


The main intuition behind our predictive approach is that knowledge of the geometry of existing occupied space can serve as a prior for generating predictions.  Prior to deep learning, the best methods of generating predictions were through {\it explicit models}, however, modeling observations and experiences can be difficult if not impossible.  Deep learning enables the ability to find hidden representations that encode prior knowledge by collecting datasets that represent experiences.  In our work, we leveraged the power of deep learning to encode prior knowledge of likely spatial structures and used this representation to generate future predictions without an explicit model.  

\begin{figure}
	\centering
   \vspace{-2.5mm} \includegraphics[width=.95\columnwidth]{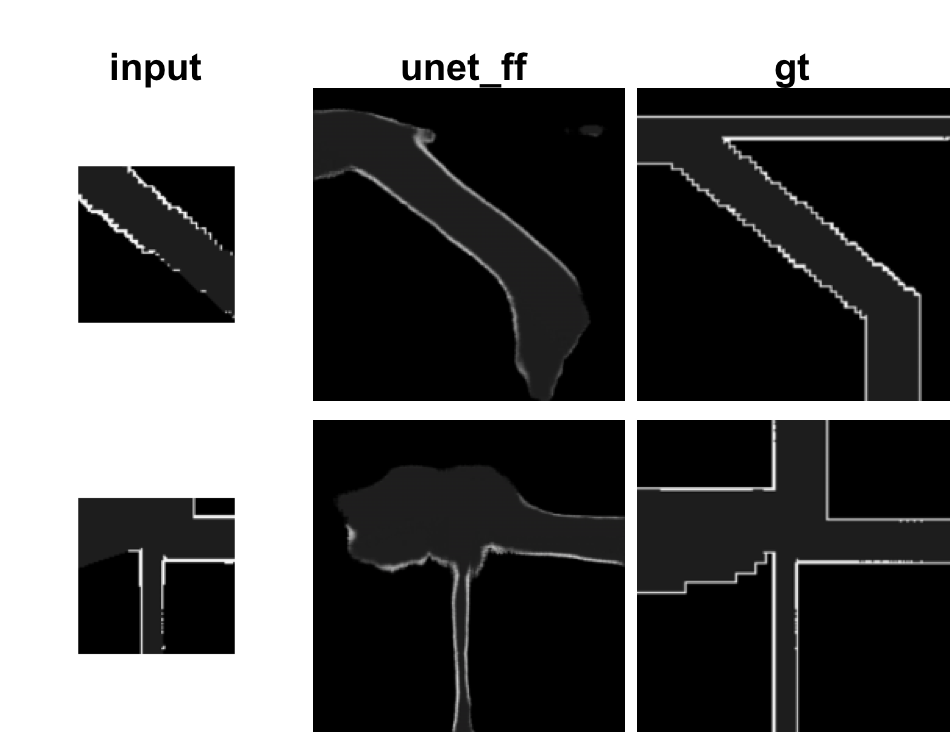}
    \caption{Failure examples when predicting 2.00x increase with the U-Net FF architecture.}
	\label{fig:bad_pred}
\end{figure}

\begin{figure}
	\centering
    \includegraphics[width=.95\columnwidth]{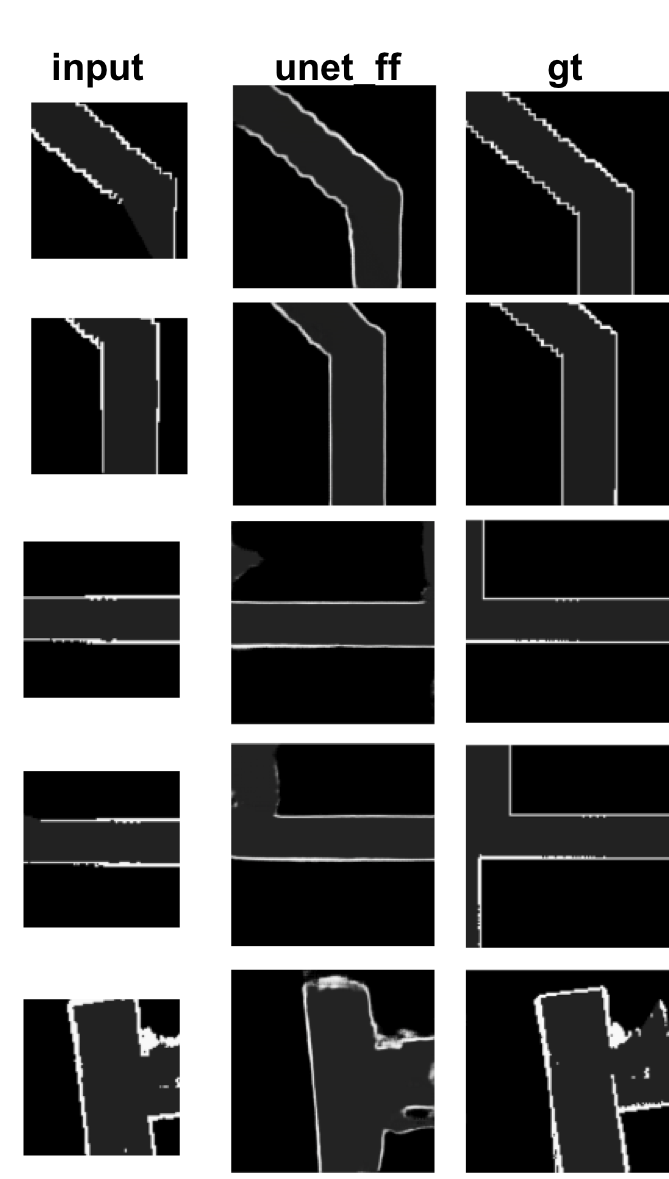}
    \caption{Random examples at 1.30x expansion using the U-Net FF architecture}
	\label{fig:random_pred}
\end{figure}

Based on the above experiments, the proposed approach is generally very stable, particularly when predicting occupancy maps representing 1.10x or 1.30x expansion increases. Considering U-Net's superior performance on the simulated data, we use it next to demonstrate the general robustness of our approach in Fig.~\ref{fig:random_pred} where we display five randomly selected images from the test dataset.  A promising benefit of our method is that with very little information, predictions can be extremely accurate as shown in Fig.~\ref{fig:good_sim_images}.  This is further evidenced by the supplementary video demonstrating accurate frame-by-frame prediction of a robot navigating a hallway in simulation.  As expected, when the predicted area of the occupancy map increases, the results exhibit more uncertainty as demonstrated by the 2.00x predictions in Fig.~\ref{fig:bad_pred}.  While falling short of the exact ground truth, these examples still contain useful information beyond the observed input, which can can be exploited by the planning algorithms.

Overall, when compared to the simulated data, the physical data performs worse quantitatively. This is likely due, to the fact that the physical data exhibits more details that are hard to predict given the simulated training data, which does not have the same level of detail (e.g., chairs, boxes, people walking through the scene). Using augmentation methods may help address this issue. 

Looking back at the physical data prediction from Fig.~\ref{fig:good_phys_images}, one notes that at a high-level the predictions are not only informative, but also all predictions are qualitatively correct as they all point to the coming of a T-like intersection. This suggests that from the perspective of the end goal of assessing navigational risks, selecting the navigation behavior, or  simply deciding on if/when to decelerate, this high level qualitative information is very useful. 

We note also that in the physical data results in Table II, there is much less quantitative difference between GAN and fully convolutional models performance, and in fact GAN seem to have a little edge qualitatively over the other methods as it is able to predict a more detailed map than the other approaches.

As demonstrated in Fig.~\ref{fig:good_phys_images}, not only can our approach be used to predict occupied space, it appears to have the beneficial effect to filter out transient obstacles found by noisy sensor readings. 
While higher resolution details might be desirable for collision avoidance, this will solved to a large extent with  the current sensor measurements in the FOV. We argue that as we expand our temporal horizon, less spatial resolution is necessary in the prediction. In this sense it would be beneficial to use alternate metrics that take this fact into account. One way to achieve this is possibly to compute SSIM at different (coarser) resolution level for more distant future time instants to characterize the ability of the prediction method to capture the future at different scales. This is left to future work.

Additional future work will also focus on improving the current methods for extending predictions and combining them with the stable results generated by the shorter horizon predictions.


\begin{table}[h]
\centering
\caption{SSIM Analysis for Physical Data}
\label{table:sa_phys}
\begin{tabular}{|c|c|c|c|} 
\hline
\textbf{Expansion} & \textbf{unet\_ff} & \textbf{resnet\_ff} & \textbf{gan} \\ [0.05ex] 
\hline\hline
1.10x & 0.511 & 0.523 & 0.489\\
\hline
1.30x & 0.517 & 0.534 & 0.524 \\
\hline
1.50x & 0.498 & 0.489 & 0.511 \\
\hline
1.70x & 0.486 & 0.504 & 0.486 \\
\hline
2.00x & 0.484 & 0.499 & 0.424 \\
\hline
\end{tabular}
\end{table}

\section{CONCLUSION}
Our long term objective is to develop risk-sensitive control algorithms capable of leveraging known obstacles in the environment as well as predicted obstacles.  In this paper, we have laid the foundation to demonstrate deep networks can be used to make predictions of occupancy maps that extend beyond the FOV of the sensor.  In our evaluation, we uncovered conditions where predictions were highly accurate and examples where the predicted results could be improved.  
As future work,  we plan to evaluate prediction mechanisms operating on raw depth data, combining visual and depth data, to develop an on-line learning policy and also to further develop risk-sensitive control policies for high speed navigation based on these predictions.

{\small
\bibliographystyle{ieee}
\bibliography{bibliography}
}

\end{document}